%% file: 0_main.tex
\documentclass[letterpaper, 10 pt, conference]{ieeeconf}
\IEEEoverridecommandlockouts
\usepackage{etextools}
\usepackage{amssymb}
\usepackage{float}
\usepackage{makeidx}
\usepackage{amsmath}
\usepackage{amsfonts}
\usepackage{bbm}
\usepackage{epsfig}
\usepackage{epsf}
\usepackage{psfrag}
\usepackage{verbatim}
\usepackage{color}
\usepackage{multirow}
\usepackage{tabularx}
\usepackage{booktabs}
\usepackage[tight,footnotesize]{subfigure}
\usepackage{array}
\usepackage{soul}
\usepackage{footnote}
\usepackage{cite}
\usepackage{dblfloatfix}

\usepackage{color, colortbl}
\usepackage{graphicx}
\graphicspath{{Figures}}
\DeclareGraphicsExtensions{.pdf,.png}
\usepackage{bigstrut}
\usepackage[english]{babel}

\usepackage{csquotes}

\usepackage[T1]{fontenc}
\usepackage{algorithm, setspace}
\usepackage{algpseudocode}
\usepackage{url}
\usepackage{multirow}
\usepackage{float}
\usepackage{xcolor}

\newcommand{\p}[1]{\smallskip \noindent \textbf{{#1}.}}
\newcommand{\eq}[1]{Equation~(\ref{eq:#1})}
\newcommand{\fig}[1]{Figure~\ref{fig:#1}}
\usepackage{balance}
\let\labelindent\relax
\usepackage{enumitem}

\usepackage[colorlinks=true,
            linkcolor=orange,
            citecolor=orange,
            urlcolor=orange,
            filecolor=orange,
            hypertexnames=false]{hyperref}
\begin{document}
\title{\LARGE
Assisting for Open-Ended Tasks: \\ Goal-Oriented Shared Autonomy as a Particle Filter
}

\author{Mengxue Fu*, Ethan Xu*, Sam Iyer-Singh*, Yinlong Dai, Michael Hagenow, and Dylan P. Losey
\thanks{*Equal contribution. This work is supported by NSF Grant $\#2337884$. Research performed at the Collaborative Robotics Lab (\href{https://collab.me.vt.edu/}{Collab}), Dept. of Mechanical Engineering, Virginia Tech, Blacksburg, VA 24061. \newline M. Hagenow leads the \href{https://wisc-rt2.github.io/}{RT$^2$ Lab}, Dept. of Computer Science, UW-Madison. \newline Corresponding author's email: \texttt{losey@vt.edu}}
}
\maketitle


\begin{abstract}

A common approach for shared autonomy blends human inputs with autonomous assistance based on the human's likely goal. However, most existing approaches assume that a static set of possible goals is known \textit{a priori}, which limits the use of such methods in unstructured assistive settings. We instead investigate how to enable shared autonomy with open-ended and dynamically changing goals. We formulate goal-oriented shared autonomy as a particle filter in which particles represent candidate human goals. Unlike conventional approaches with a fixed goal set, our transition model dynamically proposes new candidate goals as the interaction evolves, and human actions update the belief over these goals in real time. We instantiate this framework with foundation models (e.g., vision grounding and large language models) that propose context-relevant semantic goals, generate goal-conditioned assistance from low-level skill primitives, and refine those skills from human corrections. We assess our approach through a user study where 12 participants perform a variety of tabletop manipulation tasks with our method and state-of-the-art shared autonomy baselines. The results show that our particle filter-based approach reduces the amount of time users spend teleoperating the system and improves user satisfaction.
User study videos: \url{https://youtu.be/Ii26XuRqm9c}
                          
\end{abstract}


\input{1_intro}
\input{2_related}
\input{3_problem}
\input{4_theory}
\input{5_method}
\input{6_experiments}
\input{7_conclusion}

\balance
\bibliographystyle{IEEEtran}
\bibliography{Bibtex}

\end{document}

%% file: 1_intro.tex
\section{Introduction} \label{sec:intro}

\begin{figure}[t]
    \centering
    \includegraphics[width=1.0\linewidth]{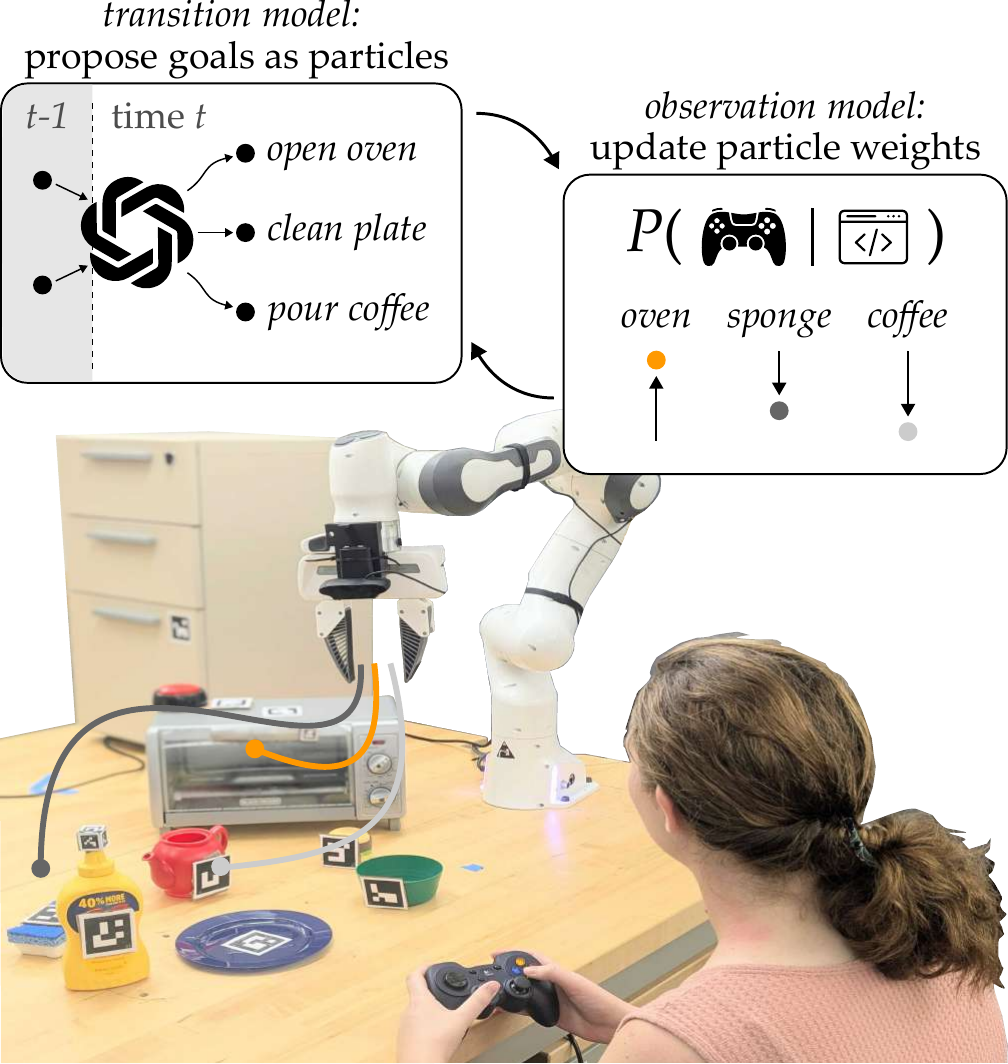}
    \caption{We frame goal-oriented shared autonomy as a \textbf{particle filter}. Given the semantic state, an LLM iteratively proposes a set of goals (i.e., particles). The particle weights are then updated based on the human's teleoperation inputs. By dynamically changing the goal set, and then generating code to assist for those goals, we enable more open-ended shared autonomy.}
    \label{fig:front}
    \vspace{-1.5em}
\end{figure}

When humans teleoperate robot arms, we often want the robot to partially automate our intended tasks.
Consider \fig{front}: by default, the human must precisely control the robot throughout the process of making a snack.
We can reduce the human's effort --- and help orchestrate the robot's low-level motions --- if we are able to infer the human's high-level \textit{goal}.
Inferring this goal is straightforward when we are given a fixed set of options (e.g., ``open the oven'' or ``clean the plate'').
But in real-world contexts these goals are open-ended and dynamic: each goal requires different assistance, and the likely goals change based on the current context.
For instance, once we grasp the sponge, there may be likely goals (such as ``use it to clean the plate'') but also several valid but unlikely goals (such as ``put it in the microwave'').
So how do robots assist operators when the robot is not only uncertain about the human's current goal, but also about the set of potential goals the human might be considering?

To answer this question we turn to \textit{shared autonomy}.
Shared autonomy provides a framework for real-time assistance in teleoperation settings \cite{hagenow2025shared}.
Ideally, a shared autonomy system can help guide the robot's motions for whatever goals the human has in mind.
One common way robots achieve this is through a \textit{predict-and-blend} approach: the robot infers the human's goal based on their actions, and then blends the human's teleoperation inputs with autonomous assistance \cite{dragan2013policy, javdani2018shared}.
Unfortunately, this standard approach to shared autonomy relies on a \textit{static} set of pre-defined goals.
Once the goals are defined, prior works apply reinforcement learning \cite{reddy2018shared, schaff2020residual} or offline demonstrations \cite{losey2022learning, karamcheti2022lila} to assist for those goals --- but the robot's assistance ultimately depends on having chosen goals that match the human's likely intents.
As we move towards real-world scenarios, it quickly becomes infeasible for robots to know all the goals (and associated motions) that the user might want to perform beforehand.

In this paper we therefore propose a shared autonomy paradigm for open-ended goals.
Unlike prior works, we never assume that the goal space is known, or that the robot has a static library of assistive policies.
Instead, we adapt the robot's assistance on the fly by recognizing that:
\begin{center} \vspace{-0.25em}
    \textit{Beyond just predicting what goal the human wants, we should also dynamically update the set of likely goals.}
    \vspace{-0.25em}
\end{center}
Applying this insight enables us to re-formulate goal-oriented shared autonomy through the lens of a particle filter, where the particles are guesses at the human's goal, and the set of particles is the goal set.
Core to achieving this open-ended shared autonomy is the ability to (a) reason about semantically meaningful goals and (b) generate assistive actions without pre-programming.
In this work, we show how foundation models --- specifically large language models (LLMs) --- can iteratively propose and refine a set of semantic goals based on the context and recent events.
We then apply neuro-symbolic tools to convert these semantic goals into grounded functions that can assist the human operator in real time.
In practice, this bi-level shared autonomy framework --- reasoning about both the goal set and the human's current intent --- can provide zero-shot assistance for previously unseen tasks.

Overall, we make the following contributions:

\p{Shared Autonomy with Dynamic Goals}
We propose a dynamic-goal shared autonomy framework, formalized as a particle filter.
Existing shared autonomy approaches can be viewed as a simplification of our framework where only the observation model is used to update the predicted goal.
We remove this constraint by incorporating a transition function that generates new sets of likely goals during interaction.

\p{Hierarchical Particle Filter Implementation}
We instantiate our transition function as an LLM that reasons over the scene and samples likely goals.
We then ground these semantic goals into code-based skills that are generated by the LLM to output assistive actions.
As the human teleoperates the robot in real time, the system infers which (if any) of the goals is the human's intent, and blends human actions with the corresponding skill.
We leverage the human's inputs as a source of feedback to improve future skill generation.

\p{Testing with Open-Ended Goals}
We perform a user study that compares our approach to state-of-the-art alternatives.
The results show that our algorithm for proposing dynamic goal sets and then inferring the human's true intent reduces user effort, even in open-ended scenarios where users are trying to perform previously unseen tasks.
We also demonstrate that continually sharing autonomy --- and keeping the human in the loop --- is essential for real-world performance: the LLM-generated skills are not perfect, and we need human interventions to fix the robot's fine-grained mistakes.

%% file: 2_related.tex
\section{Related Work} \label{sec:related}

We propose a dynamic-goal shared autonomy formalism, which is practically achieved through open-ended semantic goal inference with emergent foundation models. 
Below we review goal-oriented shared autonomy methods and the use of foundation models for goal inference.

\p{Goal-Oriented Shared Autonomy} 
In shared autonomy the human and robot jointly control the robotic system \cite{hagenow2025shared}. 
One popular paradigm is \textit{goal-oriented} shared autonomy, where the robot maintains a belief over which goals (e.g., tasks) the human is trying to accomplish \cite{dragan2013policy, javdani2018shared}. 
As the robot infers the human's goal, it provides increasing levels of assistance.
That assistance could be a modified mapping from joystick inputs to robot behaviors \cite{losey2022learning, karamcheti2022lila}, robot actions trained on previous interactions \cite{jonnavittula2024sari}, or free-space motions towards perceived goals \cite{vosa}.
However, these prior works often limit goals to geometric locations \cite{losey2022learning, vosa} or previously seen behaviors \cite{karamcheti2022lila, jonnavittula2024sari}.
By contrast, we explore settings where the goals are semantically meaningful subtasks (e.g., ``open the oven'') that dynamically change during the interaction.

\p{Implicit Goal Inference through Action-Level Assistance} 
A separate line of work does not require explicit goal inference, and instead provides assistance at the level of raw actions. 
Some methods train a copilot with reinforcement learning, correcting the human's commands from a reward signal or goal-agnostic constraints \cite{reddy2018shared, schaff2020residual, sha2026efficient}. 
Others learn an expert action distribution (for example, with diffusion or flow models) and use that distribution to guide noisy human commands, steering the resulting shared control action toward expert-like behaviors while preserving human intentions \cite{yoneda2023noise, sun2025flashback, sun2026gloves, mcmahan2024shared, wang2026disco}. 
Recent work also blends the human's commands with the actions of a frozen, pretrained policy through a computed arbitration rule \cite{zhou2026saps}. 
These methods are effective at low-level action correction and blending, which improves how an individual action is executed. 
However, they do not infer the human's intended task at a semantic level or plan over a sequence of subtasks. 
Moreover, their assistance is typically constrained by the training-time action distribution, and they may struggle with unseen tasks or behaviors outside that distribution.

\p{Semantic Goal Inference with Foundation Models} 
Foundation models have demonstrated impressive open-world perception and commonsense reasoning.
Recent studies leverage foundation models to infer the user's intent from images of short segments of teleoperation \cite{casper} or from video \cite{rahimi2026intentvlm, da2026perception}. 
They are also used to select task-aligned actions from a pretrained policy \cite{yuan2026ups}, to automatically switch joystick control modes \cite{tao2025lams}, to interpret online language corrections that refine the human's control mapping \cite{cui2023no}, and to aggregate corrections across users and tasks into general skill templates \cite{memo}. 
Closest to our approach is CASPER~\cite{casper}, which uses a vision-language model (VLM) to infer a semantic skill-object intent from the user's teleoperation history; once the intent is confirmed, the robot fully automates the corresponding skill.
However, CASPER does not share autonomy throughout the interaction, and commits to a single hypothesis rather than maintaining and refining multiple forms of assistance.

%% file: 3_problem.tex
\section{Problem Statement} \label{sec:problem}

We focus on assistive robot arms.
We consider scenarios where a human is teleoperating the robot in real time (e.g., with a joystick), and the human wants the robot arm to perform household manipulation tasks (e.g., open a cabinet).
Our overarching goal is to make it easier for the human to control the system.
Ideally, the robot recognizes what high-level task the human is trying to perform, and identifies low-level assistive actions to help automate that task.

\p{High-Level Goals}
Let $\theta_t \in \Theta$ be the human's goal at timestep $t$.
We use \textit{goal} as a general term for the human's semantically meaningful task or subtask.
Looking at our working example in \fig{front}, the goal $\theta_t$ could be to ``open the oven,'' ``clean the plate,'' or ``pour coffee.''
In theory, there is a potentially infinite set of goals the human could possibly want.
The set $\Theta$ includes possible but unlikely subtasks, e.g., ``throw the mustard'' or ``push the microwave.''

Instead of trying to define and reason over all possible goals $\Theta$, prior works in shared autonomy focus on the \textit{plausible goals} for the current scenario \cite{dragan2013policy, javdani2018shared, vosa}.
Let $\mathcal{G}_t \subset \Theta$ be a discrete space of goals at timestep $t$: the robot assumes that the human's true intent $\theta_t$ is an element of these selected goals $\mathcal{G}_t$.
We define $N_t = |\mathcal{G}_t|$ as the size of the goal space, and we instantiate $\mathcal{G}_t$ as a semantic set of goals, e.g., $\mathcal{G}_t =$ [``pick up the sponge,'' ``open the oven''].

We emphasize that the choice of $\mathcal{G}_t$ is critical for shared autonomy. 
To provide assistance the robot should infer the human's true goal $\theta_t$ during online interaction.
But because the robot only reasons over the options in $\mathcal{G}_t$, we must first ensure that $\theta_t \in \mathcal{G}_t$ and the robot maintains a set of possible goals that align with the human's likely intents.

\p{States and Actions}
The robot selects likely goals and infers the human's current intent based on the system's real-time states and actions.
Let $s = (o, x) \in \mathcal{S}$ be the system \textit{state}.
Here $o$ is a visual observation (e.g., an RGB-D image) of the scene, and $x$ is the robot's proprioceptive state (e.g., the robot's pose and the gripper's position).
Building on the raw observations and joint readings, the robot can also translate state $s$ into a templated \textit{natural language description} using object detection and localization models \cite{liu2024grounding}.
For example, ``there is an [\texttt{object label}] located at [\texttt{this position}] and oriented with [\texttt{this orientation}].''

With the state defined, we next consider the human and robot \textit{actions}.
The robot's motion is jointly controlled by a human operator and by the robot itself.
Let $a_\mathcal{H} \in \mathcal{A}$ be the action commanded by the human operator, and let $a_\mathcal{R} \in \mathcal{A}$ be the assistive action selected by the robot.
Both $a_\mathcal{H}$ and $a_\mathcal{R}$ lie in the robot's action space, e.g., they consist of end-effector velocities and commands to open or close the gripper.

\p{Converting Goals to Actions}
Inferring the human's goal is only part of the problem: the system must next convert that high-level goal $\theta$ into low-level actions $a_\mathcal{R}$ which help control the robot arm.
Let the robot have assistive policy $\pi$:
\begin{equation} \label{eq:P1}
    a_\mathcal{R} \sim \pi(\cdot \mid s, \theta_t, \rho)
\end{equation}
This policy maps the state $s$ and inferred goal $\theta_t$ into actions $a_\mathcal{R}$.
Here $\rho$ captures any prior knowledge the robot might have about this mapping, e.g., $\rho$ could be a set of demonstrations \cite{jonnavittula2024sari}, a decoder \cite{losey2022learning, karamcheti2022lila}, a pretrained VLA \cite{zhou2026saps}, or a skill library \cite{casper, memo}.
Shared autonomy typically blends the resulting assistance with human inputs to reach an overall action \cite{dragan2013policy}:
\begin{equation} \label{eq:P2}
    a = \alpha \cdot a_\mathcal{R} + (1 - \alpha) \cdot a_\mathcal{H}
\end{equation}
Here $a \in \mathcal{A}$ is the end-effector velocity and gripper command actually taken by the robot arm, and $\alpha \in [0, 1]$ is a scalar blending coefficient --- lower values of $\alpha$ mean that the human's inputs can override the robot's assistance.
At each timestep the system state transitions according to this blended action with dynamics $s_{t+1} = g(s_t, a_t)$.

\p{Summary}
To meaningfully share autonomy and assist the human, we are faced with three interconnected challenges: (a) determining the likely goals $\mathcal{G}_t$ for the current scenario, (b) inferring the human's goal $\theta_t$ from the set $\mathcal{G}_t$, and (c) mapping the selected goal into assistive actions $a_\mathcal{R}$.

%% file: 4_theory.tex
\section{Goal-Oriented Shared Autonomy \newline as a Particle Filter} \label{sec:theory}

Prior works on shared autonomy offer a variety of solutions to the challenges listed above.
However --- as we demonstrate below --- they do so while applying restrictive assumptions to the robot's goal space $\mathcal{G}_t$.
Standard approaches such as \cite{javdani2018shared, losey2022learning} treat $\mathcal{G}_t$ as a static, designer-specified list (e.g., the location of all nearby objects).
This assumption was often practically necessary since identifying relevant goals in open-ended environments was an intractable problem. 
However, recent advances in foundation models remove this constraint --- we can now enumerate common sense goals given the semantic context and interaction history.
Thus, we re-frame goal-oriented shared autonomy as a \textit{particle filter}, a non-parametric form of Bayes filtering.
Here the human's goal $\theta_t$ is treated as the hidden state, and the robot's goal set $\mathcal{G}_t$ becomes the set of particles (i.e., hypotheses) currently under consideration. 
This formulation is particularly suited to dynamic goal spaces, where particles can be added, removed, or reweighted. 
Particle filters thus provide a principled mechanism for dynamically choosing and updating goals, directly shaping both the robot's goal inference and the assistance it ultimately provides.

\subsection{Particle Filter Framework} \label{sec:t1}

A particle filter is a recursive Bayesian algorithm used to estimate the hidden state of a dynamical system \cite{gordon1993novel, arulampalam2002tutorial}.
Below we re-write the components of our shared autonomy setting within the context of a particle filter.

\p{Hidden State} 
The hidden state of our particle filter is $\theta_t$, the human's true goal at timestep $t$.
The filter maintains $N_t$ discrete hypotheses for $\theta_t$ --- each of these particles is a guess at the human's actual intent.
Overall, the particle set is: $\mathcal{G}_t = \{\theta_t^{(i)}\}_{i = 1}^{N_t}$, where $i$ indexes the particle number.
Let scalar $w_t^{(i)}$ be the likelihood associated with the $i$-th particle in $\mathcal{G}_t$.
These weights $\mathbf{w}_t=\{w_t^{(i)}\}_{i = 1}^{N_t}$ represent the probability that particle $i$ is the human's true goal; by construction, the weights are normalized so that $\sum \mathbf{w}_t=1$.

Our definitions above directly correspond to standard concepts in shared autonomy.
The particle set $\{\theta_t^{(i)}\}_{i = 1}^{N_t}$ is equivalent to the robot's discrete goal set $\mathcal{G}_t$.
The importance weights $\mathbf{w}_t$ capture the robot's belief (e.g., probability distribution) over these discrete goals, and $\mathbf{w}_0$ acts as a prior, weighting the goals at the start of the interaction.
The maximum \textit{a posteriori} estimate --- i.e., the particle with the highest weight --- serves as the robot's current estimate of the human's goal.

\p{Observation Model}
The observations (i.e., measurements) within our particle filter are the human's actions $a_\mathcal{H}$.
We use these observations to update our estimate of the human's intent; more specifically, we update the weights $\mathbf{w}_t$ across the particles.
Using Bayesian inference, we reach:
\begin{equation} \label{eq:T1}
    w^{(i)}_{t+1} \propto P\big(a_{\mathcal{H} , t}\mid s_t, \theta_t^{(i)}\big) \cdot w_t^{(i)}
\end{equation}
where the resulting weights are re-normalized such that $\sum \mathbf{w}_{t+1}=1$.
The observation model $P\big(a_{\mathcal{H} , t}\mid s_t, \theta_t^{(i)}\big)$ tells us how likely it is to observe human input $a_{\mathcal{H},t}$ given that the system is currently in state $s_t$ with the hypothesized goal $\theta_t^{(i)}$.
For example, this probability could be based on the cosine alignment between the human's input $a_\mathcal{H}$ and the assistive action $a_\mathcal{R}$ for the hypothesized goal.
We emphasize that updating the particle weights in \eq{T1} is functionally identical to other shared autonomy works that apply Bayesian inference to obtain a belief across goals \cite{dragan2013policy,javdani2018shared,losey2022learning}.

\begin{figure*}[t]
    \centering
    \includegraphics[width=1\linewidth]{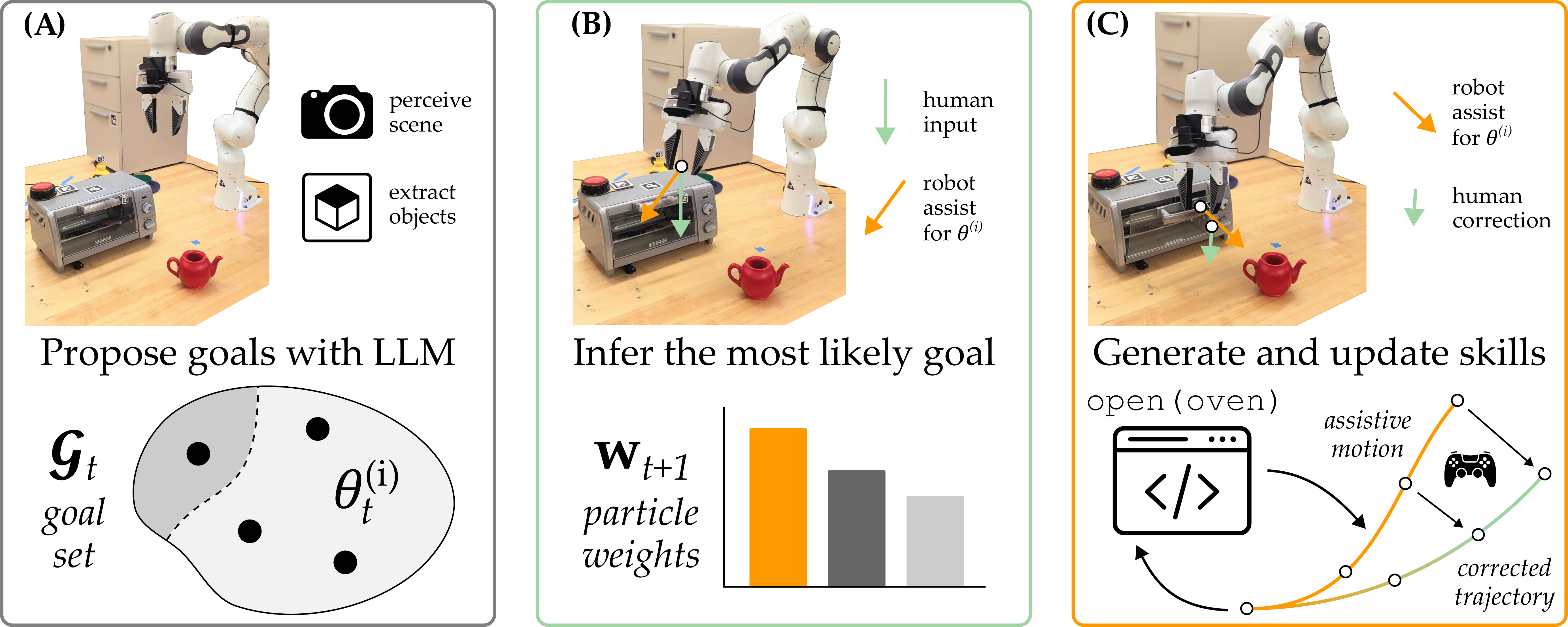}
    \caption{Our proposed approach for goal-oriented shared autonomy. (A) Given an image of the scene, we extract objects and fill in a natural language state description. This state description is then provided to an LLM, which functions as the \textit{transition model} and proposes semantic goals $\mathcal{G}_t$. The number of proposed goals is adapted based on coverage. (B) The weight of each goal in $\mathcal{G}_t$ is updated using Bayesian inference based on our \textit{observation model}, the cosine alignment between the goal-specific assistance and the human's joystick inputs. (C) The robot provides assistance for the most likely goal. To find a policy that assists for that goal, an LLM dynamically generates \textit{skills} (i.e., code blocks) which are executed by the robot. When the human intervenes during shared autonomy and corrects the robot's behavior, we store that corrected trajectory and use it to condition future skill generation.}
    \label{fig:method}
    \vspace{-1.5em}
\end{figure*}

\p{Transition Model}
The final component of our particle filter is the transition model, which predicts how the hidden state $\theta_t$ evolves over time.
More practically: the transition model generates the next set of particles (i.e., hypotheses) to estimate the human's goal.
We define our transition model $f$ as a function of the current state, particle set, and weights:
\begin{equation} \label{eq:T2}
    \mathcal{G}_{t+1} \sim f( \cdot \mid s_t, \mathcal{G}_t, \mathbf{w}_t)
\end{equation}
Note that the next particle set $\mathcal{G}_{t+1}$ may change in size as compared to $\mathcal{G}_{t}$, and the model could be conditioned on a particle history that goes beyond $\mathcal{G}_t$.
Intuitively, we expect the transition model to output particles that are aligned to likely hypotheses, while accounting for how the state $s$ changes during the interaction.
In terms of our example: if the human has teleoperated the robot to open the oven, the next particle set might include ``put food in the oven.''

\subsection{Special Case: Static Goal Set} \label{sec:t2}

The hidden state and observation model from Section~\ref{sec:t1} have direct equivalents in existing shared autonomy methods.
However, the transition model is often constrained.
We can view predict-and-blend approaches \cite{losey2022learning, javdani2018shared, dragan2013policy} as instantiations of our particle filter with the reduced transition model:
\begin{equation} \label{eq:T3}
    \mathcal{G} \sim f(\cdot \mid \Theta, \mathbf{w}_0)
\end{equation}
This transition model inputs the space of all possible goals $\Theta$ and some initial weights $\mathbf{w}_0$.
The model is called once --- at the start of the interaction --- and outputs a static set of plausible goals $\mathcal{G}$.
In practice, the user-specified weights $\mathbf{w}_0$ determine which goals are selected and the relative likelihood of these goals (i.e., the prior).
\textit{The goals are not updated throughout the task.}
Instead of repeatedly calling the transition model like a typical particle filter, this simplification only uses the observation model in \eq{T1} to update the weights $\mathbf{w}_t$ and determine which of the goals in $\mathcal{G}$ is best aligned with the human's teleoperation inputs. 

\subsection{General Case: Dynamic Goal Set} \label{sec:t3}

If we remove this constraint, we reach a particle filter for shared autonomy that iteratively calls the transition model to propose new sets of likely goals.
The update steps include:
\begin{itemize}
    \item Using the observation model to obtain new weights $\mathbf{w}_{t+1}$ over particles $\mathcal{G}_t$ based on human inputs.
    \item Using the transition model to propose new particles $\mathcal{G}_{t+1}$ based on the context and likely weights.    
\end{itemize}
The most likely particle in $\mathcal{G}_t$ serves as our current guess at the human's goal.
These steps can be performed at the same or different frequencies, and the precise forms of the observation and transition functions can vary between implementations.
We want to highlight that --- by recognizing that shared autonomy can be framed as a particle filter, and by lifting the restriction on the transition function --- we enable a more general class of shared autonomy approach that draws from particle filter methods.
In particular, particle filters provide principled techniques for proposing new particles $\mathcal{G}_{t+1}$ \cite{gordon1993novel, arulampalam2002tutorial} and adapting the number of particles $N_{t+1}$ \cite{fox2003adapting}.
We will leverage these concepts in the following section to derive our shared autonomy algorithm for dynamic goals.

%% file: 5_method.tex
\section{Open-Ended Shared Autonomy} \label{sec:method}

Here we present an instantiation of the particle filter formalism from Section~\ref{sec:theory}.
Our algorithm has three main parts (see \fig{method}).
First, we parse the state into a natural language description and leverage an off-the-shelf LLM to serve as the \textit{transition model}: this LLM is prompted to propose context-relevant goals, and the number of goals is then adapted to obtain the next particle set $\mathcal{G}_{t}$.
Second, we infer the most likely goal $\theta \in \mathcal{G}_t$ using a real-time \textit{observation model}.
This observation model considers how the human's teleoperation inputs align with the motions for each possible goal. 
If the human's inputs do not align with any proposed goals, the robot returns full control to the human.
Finally, we maintain and update an assistive \textit{skill library} that includes parameterized functions for individual goals, e.g., \texttt{Open(Oven)}. 
The robot retrieves and leverages these functions within its policy $\pi$ to provide goal-specific assistive actions.
The skills are generated and iteratively updated using an LLM conditioned on how the human has previously teleoperated the robot for the given goal.

\subsection{Transition Model: Proposing a Goal Set} \label{sec:M1}

We first present our transition model from \eq{T2}.
This transition model is implemented as an LLM which updates at a lower frequency than the real-time controller.
At timesteps when an update occurs, the LLM receives a natural language description of the state $s_t$ and the history of completed and rejected goals $h = \{\theta_1, \ldots, \theta_{t-1}\}$.
Based on this context, the LLM is prompted to generate a set of $N_t$ independent candidate particles, $\mathcal{G}_t=\{\theta_t^{(i)}\}_{i=1}^{N_t}\subset\Theta$.
We normalize the structure of each generated particle as
$\theta^{(i)} = \{\theta^{(i)}_{\mathrm{act}}, \theta^{(i)}_{\mathrm{obj}}\}$. 
Specifically, every $\theta^{(i)} \in \mathcal{G}_t$ follows the format \texttt{ACTION(OBJECT)}.

The number of plausible goals $N_t$ depends on the current context. 
Consider \fig{front}: at the start of the interaction there are diverse tasks that could be completed, but once the user controls the robot arm to open the oven, their next goal narrows down to a few likely options.
We therefore adapt the number of particles $N_t$ using a coverage-based stopping criterion driven by the Good-Turing estimator \cite{good1953}.
At each transition step, parallel LLM calls generate a batch of stochastic, independently sampled task proposals.
These proposals are generated in \textit{parallel} for computational efficiency, but then processed \textit{sequentially} for adaptive particle filtering.
When iterating through the proposed goals, the robot groups these proposals into semantically equivalent bins. 
Using LLM filtering, two goal predictions are assigned the same bin when they describe the same underlying action $\theta_{\mathrm{act}}$ and target object $\theta_{\mathrm{obj}}$ despite minor differences in wording.

As the predictions are processed, we apply the Good-Turing coverage estimator \cite{good1953} to prune the smaller bins.
Let $n_t$ denote the total number of goal predictions and let $f_{1,t}$ denote the number of bins containing exactly one proposal.
The estimated \textit{coverage} $\widehat{C}_t$ (i.e., the probability that a new proposal falls in a previously seen bin) is defined as:
\begin{equation} \label{eq:M1}
    \widehat{C}_t = 1-\frac{f_{1,t}}{n_t}.
\end{equation}
A goal that was only proposed once signals a tail of the distribution that has not been explored.
Hence, we sequentially process the sampled goals until $\widehat{C}_t \geq 0.95$, at which point the evaluation stops and each bin becomes a particle in the current set $\mathcal{G}_t$ so that $N_t = |\mathcal{G}_t|$.
In practice, this termination removes goals that were only proposed once, and retains goals that the LLM has proposed multiple times in parallel calls.
The number of different goals $N_t$ intuitively increases when the LLM outputs a wider range of possible particles.

We leverage the size of the bins to initialize the particle weights $\mathbf{w}_t$.
Let $|f_t| \leq n_t$ be the number of proposed goals that fall in the $N_t$ bins: the initial weight of goal $i$ is equivalent to the number of times that goal was proposed by the LLM, normalized by the total number of proposals $|f_t|$ across the $N_t$ accepted goals.

\subsection{Observation Model: Inferring the Most Likely Goal} \label{sec:M2}

Once the goal set $\mathcal{G}_t$ is obtained, we next try to infer the user's intent $\theta \in \mathcal{G}_t$.
In practice, we perform maximum \textit{a posteriori} (MAP) estimation, where the particle $\theta_t^{(i)}$ with the highest weight $w_t^{(i)}$ is assumed to be the human's true goal.
Following \eq{T1}, we leverage an observation model to update the weights $\mathbf{w}_t$ and determine the most likely particle.
We note that this inference occurs in real time, and updates at each timestep $t$ based on the human's teleoperation inputs.
We also recognize that the human's true intent may not be captured by any of the goals in $\mathcal{G}_t$, and when this occurs we should not interfere with the human's actions.

\p{Cosine Alignment}
Referring to \eq{T1}, we leverage cosine similarity as the basis for our observation model:
\begin{equation} \label{eq:M2}
   P\big(a_\mathcal{H}\mid s, \theta\big) \propto \exp{\big[\beta \cdot (\text{sim}(a_\mathcal{H}, a_\mathcal{R}) + 1)\big]}
\end{equation}
where $\text{sim}(\cdot , \cdot)$ is the cosine similarity between two vectors and $\beta \geq 0$ is the temperature constant for inference (increasing hyperparameter $\beta$ leads to faster weight updates).
\eq{M2} focuses on the alignment between the human's teleoperation input $a_\mathcal{H}$ and the robot's assistive actions $a_\mathcal{R}$.
As highlighted in \eq{P1}, these assistive actions are conditioned on the current state $s$ and the inferred goal $\theta$.
We can therefore (a) iterate through the proposed goals $\theta \in \mathcal{G}_t$ and use the robot's assistive policy to compute their respective actions $a_\mathcal{R}$ at state $s$.
We then (b) calculate $P\big(a_\mathcal{H}\mid s, \theta\big)$ based on the alignment between $a_\mathcal{H}$ and each of these potential robot actions.
During implementation we averaged $a_\mathcal{H}$ over $1$ second of inputs to obtain a smoother input signal for calculating \eq{M2}.

We will discuss the assistive policy $a_\mathcal{R} \sim \pi(\cdot \mid s, \theta, \rho)$ in the following subsection.
For now, we emphasize that actions $a_\mathcal{R}$ are designed to complete the corresponding goal $\theta$.
These actions are not necessarily a straight line to the object --- for instance, if the robot thinks the goal is \texttt{Open(Oven)}, its actions align the robot's gripper with the oven door and then pull this door open.
However, in cases where the robot is unsure how to assist for a goal $\theta \in \mathcal{G}_t$, we simplify $a_\mathcal{R}$ and assume it is a straight line from $s$ towards $\theta_{\mathrm{obj}}$.

\p{Particle Weight Update}
With our observation model instantiated, we can now leverage it to update the weights (i.e., the likelihood) of each particle.
Here we follow standard Bayesian inference to get $\mathbf{w}_{t+1}$, but with two edge cases.
First, we apply an upper bound to the particle weights such that $w_{t+1}^{(i)} \leq 0.95$.
This practically prevents the particles from collapsing and the robot becoming overconfident in a single goal.
Second, we apply a threshold on the cosine similarity in \eq{M2}: if the alignment between the human's inputs and every goal-specific action is less than $0$, then the robot does not select any particle for assistance.
Intuitively, this captures scenarios where either the robot has not proposed the human's goal, or the robot is providing incorrect assistance for that goal.
The transition model from Section~\ref{sec:M1} is running in parallel to this inference: new goal sets $\mathcal{G}_{t+1}$ are iteratively being proposed, and these proposals are conditioned on the history of selected particles.
For example, if the robot observes that none of the goals in $\mathcal{G}_t$ explain the human's inputs, then this feedback is used by the LLM to condition its next proposal $\mathcal{G}_{t+1}$.

\subsection{Skill Library: Converting Goals to Assistance} \label{sec:M3}

\begin{figure*}[t]
    \centering
    \includegraphics[width=1\linewidth]{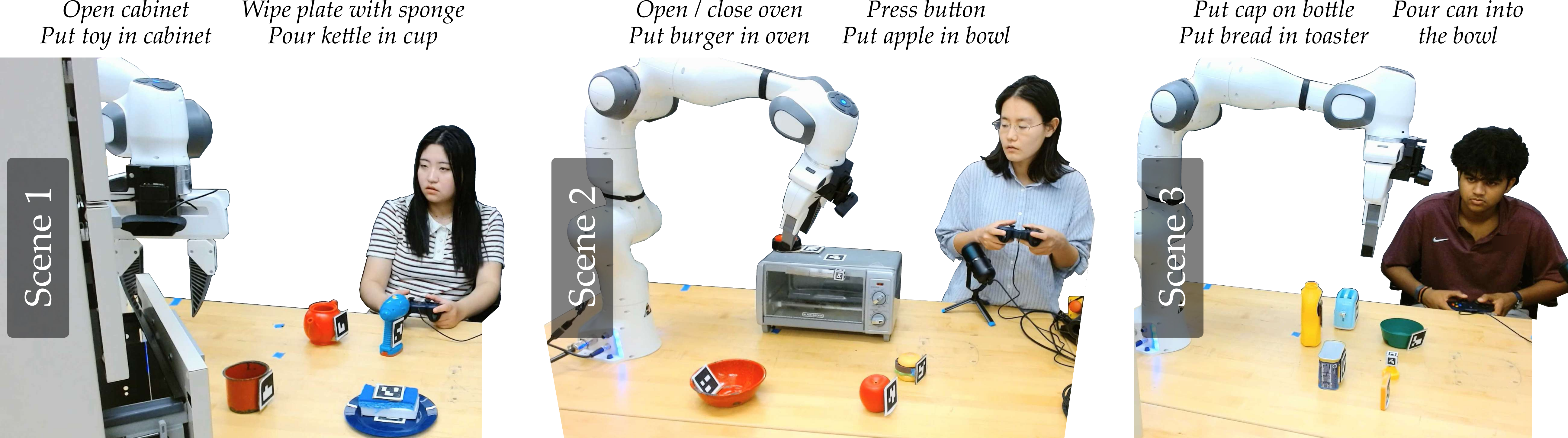}
    \caption{User study tasks and scenes. In all scenes the goals were open-ended --- the robot was never told what tasks the human might want to perform. Scenes 1 and 2 were \textit{in-distribution}: \textbf{SAPS} was fine-tuned on the tasks, and \textbf{CASPER} and \textbf{Ours} were given a library of goal-relevant skills. Scene 3 was \textit{out-of-distribution}, with no fine-tuning or skill library. Users completed Scene 3 three times to assess how \textbf{Ours} generated and corrected assistance.}
    \label{fig:tasks}
    \vspace{-1.5em}
\end{figure*}

Using the transition model from Section~\ref{sec:M1}, we first obtain a dynamic goal set $\mathcal{G}_t$.
Next, applying the observation model from Section~\ref{sec:M2}, we identify the most likely particle $\theta_t \in \mathcal{G}_t$ --- or return that no particle is aligned with the human's inputs.
Consider the case where the robot does infer a goal $\theta_t \in \mathcal{G}_t$: how can the system \textit{assist} for that goal? 
Below we describe our implementation of the assistive policy $\pi$ from \eq{P1}.
Our approach is neuro-symbolic: we leverage the LLM to generate and update a skill library $\rho$ that contains goal-specific code blocks.
In general, our method follows recent works like Code as Policies \cite{liang2023code}.
The main difference stems from our application: because we have a human in the loop, we can leverage the human's teleoperation inputs to continuously refine the robot's generated skills.

\p{Retrieving and Generating Skills}
Let $\rho = \{S_1, S_2, \ldots \}$ be a skill library and let $\mathcal{S}_k$ be the $k$-th skill in that library.
Each skill $\mathcal{S}_k$ has a semantic label of the form \texttt{Action(Object)}.
This labeling convention matches our semantic goals $\theta^{(i)} = \{\theta^{(i)}_{\mathrm{act}}, \theta^{(i)}_{\mathrm{obj}}\}$.
Hence, it is straightforward for the system to retrieve relevant skills: given a proposed goal $\theta$, the policy iterates through the library $\rho$ to see if there is a skill with the same action and object label.

The skills themselves are parameterized functions that are dynamically generated by an LLM \cite{liang2023code}.
This skill generation is conditioned on the robot's library of low-level functions (e.g., \texttt{move\_to\_pose}), the natural language task description given by the goal (e.g., ``open the oven''), and the most recent human correction (discussed below).
At run time, the robot uses the semantic decomposition of state $s_t$ to identify relevant parameters and apply them to the skill.
For instance, when executing \texttt{Open(Oven)}, the robot inputs its current pose, the pose of the oven, and the angle of its door.
Because shared autonomy requires real-time execution, we do not rewrite the skills at run time.
Instead, the matching skill (if one exists) is retrieved from the library $\rho$ and directly executed with relevant parameters from $s_t$.

The actions $a_\mathcal{R}$ output by the skill are blended with the human's inputs $a_\mathcal{H}$ following \eq{P2}.
By default the blending coefficient is $\alpha = 0$ so that the human fully controls the system.
When the robot's belief $w_t^{(i)}$ in a particle $\theta_t^{(i)}$ exceeds $0.8$, then we set $\alpha = 0.1$ and provide assistance for the likely goal.
We purposely keep the level of assistance low so the human can always override the system.
Gripper commands are treated as binary --- either open or close --- and these gripper commands are fully specified by the user.

\p{Correcting Skills}
So far our skill generation is \textit{open-loop}: the LLM writes code it thinks will help for the given goals.
But we know these functions can fall short \cite{memo}, and the robot may not have the skills $\mathcal{S} \in \rho$ it needs for a new goal $\theta$.
To address this gap we leverage the inherent collaboration of shared autonomy.
The robot's assistance does not need to be perfect: instead, the human operator can always guide the system through new tasks and make corrections to existing behaviors, providing context for skill improvement.

Our approach dynamically grows and refines the robot's skill library $\rho$ based on the human's teleoperation inputs.
Let $\xi_\mathcal{H}$ be the trajectory of states and human inputs for a completed goal $\theta$.
The robot automatically records these trajectories throughout the interaction, and then compresses them into a compact sequence of waypoints $\hat{\xi}_\mathcal{H}$ \cite{muckell2014compression}.
This results in pairs $(\theta, \hat{\xi}_\mathcal{H})$ which include the semantic task label and the human's correction for that task.
We apply these task-correction pairs $(\theta, \hat{\xi}_\mathcal{H})$ to \textit{condition} future skill generation.
If the robot is completing a goal for which it does not have an existing skill (e.g., a new task), then the LLM generates its \texttt{Action(Object)} skill conditioned on the example $(\theta, \hat{\xi}_\mathcal{H})$.
Alternatively, if the robot is performing a goal for which it has the relevant skills, we refine those skills using $(\theta, \hat{\xi}_\mathcal{H})$ and the code the robot previously executed.
In summary: we use LLMs to write assistive skills, and these skills can be augmented or refined based on the human's teleoperation inputs for previously seen goals.

%% file: 6_experiments.tex
\section{User Study} \label{sec:experiments}

To assess how our particle filter-based approach compares to state-of-the-art alternatives, we conducted a user study where $N=12$ participants shared autonomy with a robot arm.
Each participant teleoperated the robot to complete the tasks shown in \fig{tasks}.
For some of these tasks we initialized the robot with a relevant skill library (\textit{in-distribution}), while for other tasks the robot had to build and correct skills over repeated interactions (\textit{out-of-distribution}).
\textit{None of the tasks were known by the system \textit{a priori}, and the robot had to determine the user's potential goals and provide assistance on-the-fly.}
Videos from our user study can be found here: \url{https://youtu.be/Ii26XuRqm9c}

\p{Experimental Setup}
Participants sat next to the table and controlled the robot with a joystick.
We designed three scenes for interaction, each of which contained multiple objects and tasks.
This diversity was an important challenge for our method: when there are only a few objects in the scene, there are fewer plausible goals for the system to propose.

To prevent confounding factors we used ArUco markers to label the objects.
As we show in the supplemental video, our method also works \textit{without} any ArUco markers by leveraging open vocabulary classifiers.
However, during our user study we did not want the classifier performance to affect our method or the baselines, and so this step was omitted.
The robot observed its scene using a fixed base OAK-D camera and a Logitech C920 webcam attached to its end-effector.

\p{Independent Variables}
We compared four methods for providing assistance: direct teleoperation (\textbf{Teleop}), \textbf{SAPS} \cite{zhou2026saps}, \textbf{CASPER} \cite{casper}, and \textbf{Ours}.
SAPS uses a VLA to select assistive actions.
Following \cite{zhou2026saps}, we used $\pi_{0.5}$ fine-tuned on the DROID dataset as well as an additional 10 demonstrations for each \textit{in-distribution} task.
Users directly conveyed their goal to \textbf{SAPS} through verbal inputs (e.g., stating ``open the cabinet'') --- this privileged information was not available to \textbf{Ours}.
Like \textbf{Ours}, \textbf{CASPER} uses a foundation model to anticipate likely goals and relies on a skill library.
We initialized \textbf{CASPER} and \textbf{Ours} with the same skill library: this library only included goals relevant for Scenes 1 and 2.
Both \textbf{CASPER} and \textbf{Ours} performed inference in real time with GPT-OSS-120B.

\p{Participants and Procedure}
We recruited $12$ participants from the campus community ($2$ female, average age $22.3 \pm 2.1$, and $6$ with robot experience).
All participants provided informed written consent following university guidelines.

At the start of the experiment each participant was given 5 minutes to practice directly teleoperating the robot on tasks from Scene 1 and Scene 2.
The participant then completed each task with a given shared autonomy algorithm.
In Scenes 1 and 2 the participants only completed the tasks once, while in Scene 3 users repeated the tasks a total of three times with \textbf{Ours} (to assess skill library improvement).
The order of the methods was counterbalanced using a Latin square design.

\begin{figure*}[t]
    \centering
    \includegraphics[width=1.0\linewidth]{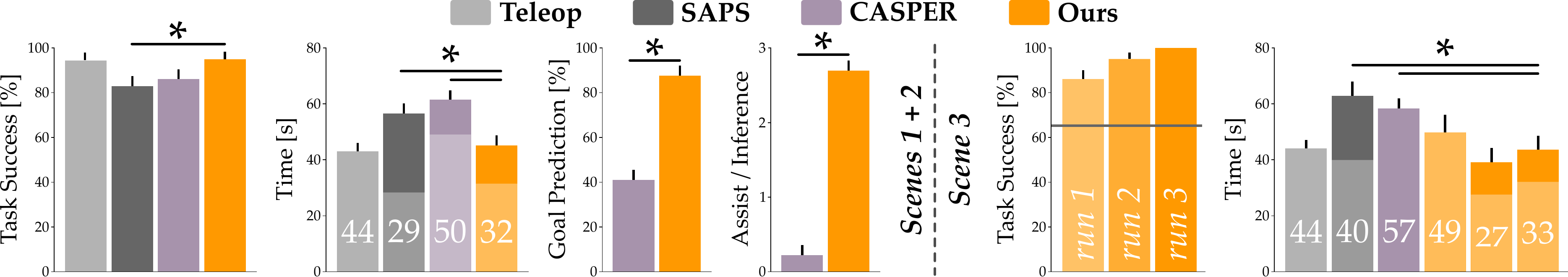}
    \caption{(Left) Objective results across \textit{in-distribution} Scenes 1 and 2. (Right) Objective results for \textit{out-of-distribution} Scene 3. For \textit{Time}, the solid bar shows the total time to complete a task, and the shaded region (with the listed number) is the amount of time the human spent teleoperating the robot. To examine the effect of assistance type, we conducted repeated measures ANOVAs followed by \textit{post hoc} comparisons using Bonferroni-corrected t-tests. This same analysis procedure was followed in \fig{subjective}. Error bars show standard error, and an $*$ denotes statistical significance ($p<.05$). }
    \label{fig:objective}
    \vspace{-1.5em}
\end{figure*}

\begin{figure}[t]
    \centering
    \includegraphics[width=1.0\linewidth]{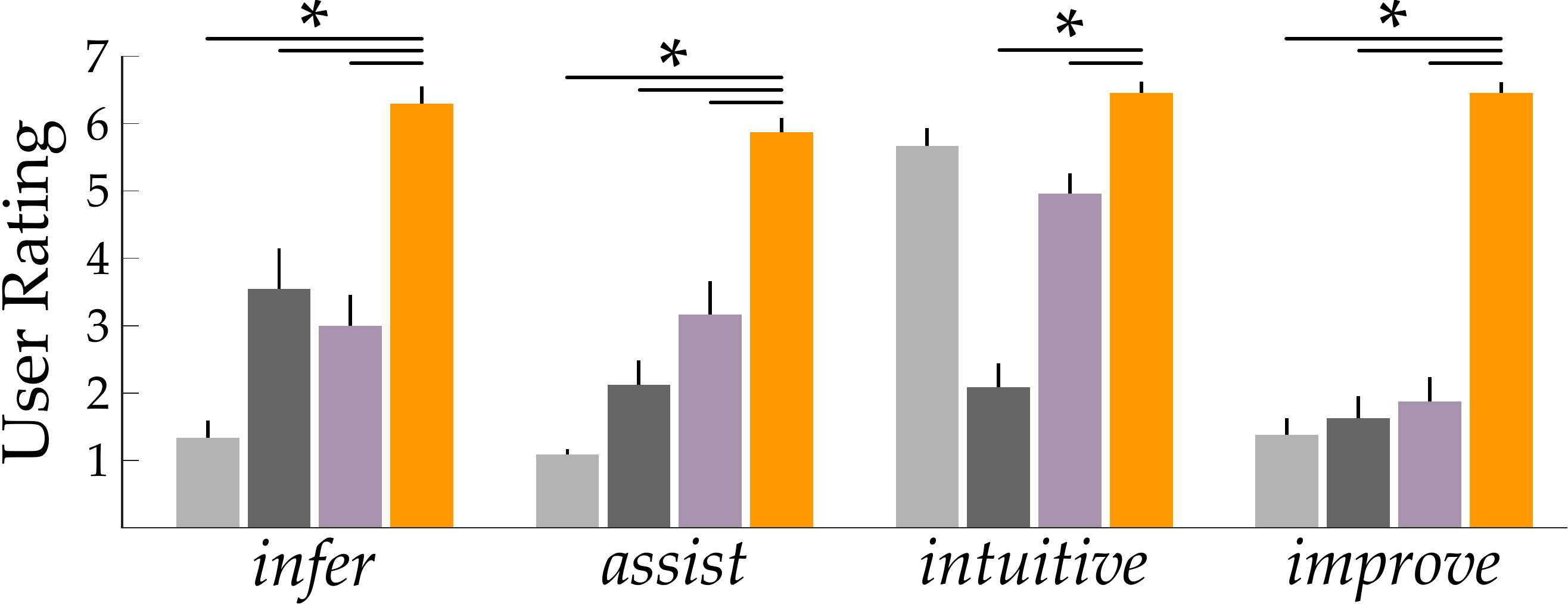}
    \caption{Subjective results. Error bars show standard error, and an $*$ denotes statistical significance ($p<.05$). When asked to select their preferred method, 11/12 users chose \textbf{Ours}, and the remaining user selected \textbf{Teleop}.}
    \label{fig:subjective}
    \vspace{-1.5em}
\end{figure}

\p{Dependent Variables}
We measured \textit{Task Success}, the percentage of trials where users completed the task in less than 90 seconds.
We also measured \textit{Time}, including both the total time taken to complete the task as well as the user's control time.
The user's control time is calculated as the number of timesteps where the human is applying inputs to the joystick or pressing buttons.
Specifically for \textbf{CASPER} and \textbf{Ours} we recorded \textit{Goal Prediction}, the percentage at which the system correctly inferred the user's goal, and \textit{Assist/Inference}, the amount of time spent assisting the human divided by the time spent by the LLM trying to infer the goal(s).
After each method we administered a 7-point Likert scale survey asking about the user's experience.
Items were organized along four scales: how effectively the robot \textit{inferred} the human's goals, how much \textit{assistance} the user felt the robot provided, how \textit{intuitive} it was to share autonomy, and how much the system \textit{improved} over time.

\p{Results}
Our objective results are summarized in \fig{objective}.
For \textit{in-distribution} tasks where the robot was given relevant fine-tuning or skills, we found that \textbf{Ours} had a higher success rate and lower total time than \textbf{SAPS} and \textbf{CASPER}.
Looking specifically at \textbf{CASPER}, we find that \textbf{Ours} is able to more accurately infer the human's goal, and spends more of its time providing assistance as compared to performing inference.
The total time for \textbf{CASPER} increased because it often inferred the wrong task, forcing users to override the system until the correct goal was predicted.
For \textit{out-of-distribution} tasks we found that \textbf{Ours} improved its assistance over repeated interactions.
While \textbf{SAPS} was successful roughly 67\% of the time, \textbf{Ours} reached 100\% success rates after three interactions.
In the first interaction the user had to teleoperate the robot through the entire task (since the robot had an empty skill library).
By conditioning on the human's examples, \textbf{Ours} generated assistive skills that reduced the human's teleoperation burden and increased success in runs 2 and 3.
Note that the time for \textbf{CASPER} is higher \textit{out-of-distribution} because it still tried to infer the human's goal, but it lacked the ability to update its skill library for any new goals.
In our experiments \textbf{Ours} and \textbf{Teleop} had similar task success and completion times, but with \textbf{Ours} the amount of time users spent teleoperating the robot was reduced.

Our subjective results are summarized in \fig{subjective}.
Users indicated \textbf{Ours} was better at inferring their goal, providing assistance, and dynamically improving as compared to all baselines.
User responses also suggest that \textbf{Ours} was more intuitive than \textbf{SAPS} and \textbf{CASPER}, and roughly as intuitive as just performing direct teleoperation.
When explaining their choices, users mentioned that \textbf{SAPS} ``often resisted their inputs,'' while \textbf{CASPER} ``kept interrupting by predicting the wrong goal.''
Overall, $11/12$ participants ranked \textbf{Ours} as their most preferred method, and $8/12$ participants ranked \textbf{SAPS} as their least preferred method.

%% file: 7_conclusion.tex
\section{Conclusion}

We explore goal-oriented shared autonomy when the robot is not only uncertain about the human's goal, but also about what goals are plausible.
We theoretically frame this problem as a particle filter, where each goal is a particle, and new particles are proposed and weighted during interaction.
Our resulting algorithm instantiates the transition model as an LLM that reasons over the scene and samples semantic goals.
We then apply a neuro-symbolic pipeline to ground these semantic goals into low-level, parameterized skills that can be generated on the fly and corrected by human inputs.
During testing we show that this framework enables shared autonomy for open-ended goals: the robot is able to recognize what the human wants and synthesize partial autonomy, even when faced with previously unseen scenarios. 